\documentclass[conference]{IEEEtran}
\IEEEoverridecommandlockouts

\usepackage{times}
\usepackage{soul}
\usepackage[utf8]{inputenc}
\usepackage{graphicx}
\usepackage{amsmath}
\usepackage{booktabs}
\usepackage{caption}
\usepackage{amsmath}
\usepackage{mathtools}
\usepackage{standalone}
\usepackage{siunitx}
\usepackage{booktabs}
\usepackage{subcaption}
\usepackage{url}
\usepackage[capitalize]{cleveref}
\usepackage{etoolbox}
\usepackage{tikz}
\usetikzlibrary{shapes,intersections,calc,shadings}

\DeclareMathOperator*{\argmin}{arg\,min}
\DeclareMathOperator*{\dist}{dist}

\newcommand{\anonymizeurl}[1]{%
#1%
}

\title{Learning Embeddings of Directed Networks with Text-Associated
  Nodes---with Application in Software Package Dependency Networks}

\author{

 \IEEEauthorblockN{%
   Kexuan Sun\IEEEauthorrefmark{1}\quad
   Shudan Zhong\IEEEauthorrefmark{2}\quad
   Hong Xu\IEEEauthorrefmark{1}\textsuperscript{\textsection}
  }%
  \IEEEauthorblockA{\IEEEauthorrefmark{1} University of Southern California, Los Angeles, California 90089, USA}%
  \IEEEauthorblockA{\IEEEauthorrefmark{2} University of California, Berkeley, California 94720, USA}
  kexuansu@usc.edu\quad sdz@berkeley.edu\quad hongx@usc.edu
  }
  
\date{}

\begin{document}
\maketitle
\begingroup\renewcommand\thefootnote{\textsection}
\footnotetext{Now at International Business Machines (IBM) Corp.}
\endgroup

\begin{abstract} A network embedding consists of a vector representation for
each node in the network. Its usefulness has been shown in many real-world
application domains, such as social networks and web networks. Directed networks
with text associated with each node, such as software package dependency
networks, are commonplace. However, to the best of our knowledge, their
embeddings have hitherto not been specifically studied. In this paper, we
propose PCTADW-1 and PCTADW-2, two algorithms based on neural networks that
learn embeddings of directed networks with text associated with each node. We
create two new node-labeled such networks: The package dependency networks in
two popular GNU/Linux distributions, Debian and Fedora. We experimentally
demonstrate that the embeddings produced by our algorithms resulted in node
classification with better quality than those of various baselines on these two
networks. We observe that there exist systematic presence of analogies (similar
to those in word embeddings) in the network embeddings of software package
dependency networks. To the best of our knowledge, this is the first time that
such systematic presence of analogies is observed in network and document
embeddings. We further demonstrate that these network embeddings can be novelly used for better understanding software attributes, such as the development process and user interface of software, etc.
\end{abstract}

\section{Introduction}

Machine learning has a long history of being applied to networks for
multifarious tasks, such as network classification~\cite{snbgge08}, prediction
of protein binding~\cite{adwf15}, etc. Thanks to the advancement of technologies
such as the Internet and database management systems, the amount of data that
are available for machine learning algorithms have been growing tremendously
over the past decade. Among these datasets, a huge fraction can be modeled as
networks, such as web networks, brain networks, citation networks, street
networks, etc.~\cite{xskk18}. Therefore, improving machine learning algorithms
on networks has become even more important.

However, due to the discrete and sparse nature of networks, it is often
difficult to apply machine learning directly to them. To resolve this issue, one
major school of thought to approach networks using machine learning is via
network embeddings~\cite{gf18}. A network embedding consists of a real
number-valued Euclidean vector representation for each node in the network.
These vectors can then be fed into machine learning algorithms for various
classification and regression tasks.

In recent years, there have been dramatic advancements in learning network
embeddings, such as DeepWalk~\cite{pas14} (and its variants using personalized graph or node context distributions~\cite{apaa2018,hhhs2020}), LINE~\cite{tqwzym15},
node2vec~\cite{gl16}, DNGR~\cite{clx16}, 
metapath2vec~\cite{dcs17}, 
M-NMF~\cite{wcwpzy17},
PRUNE~\cite{lhcyl17}, 
GraphSAGE~\cite{hyl17}, LANE~\cite{hlh17},
RSDNE~\cite{wywwwl18}, ANE~\cite{dai18}, ATP~\cite{jba19} and SIDE~\cite{kplk18}. Most of
these works, however, focus on the structure of the networks alone and do not
take data associated with nodes into account. But, in reality, there exist a
huge amount of networks in which each node is associated with text data (a
document), such as citation networks with titles and abstracts of articles, web
networks with contents of web pages, etc.

To ameliorate this issue, \cite{ylzsc15} proposed \textit{text-associated
DeepWalk} (TADW), a network embedding learning algorithm that leverages the
network and documents associated with nodes. Despite the presence of
``DeepWalk'' in its name, it does not directly use random walks. Instead, it is
based on matrix factorization of adjacency matrices and thus bears scalability
issues: Its space requirement scales quadratically with respect to the number of
nodes. Therefore, developing an effective and scalable network embedding
learning algorithm for networks with text-associated nodes remains imperative.
Another related work is Paper2vec~\cite{sv17}, which, however, is tailored for
research paper citation networks.

One important kind of network with text-associated nodes is \textit{software
package dependency networks} (SPDN). They play essential roles in modern
software package management systems. For example, in most cases, when a user
installs a software package on a modern GNU/Linux distribution via its software
package management system (which is the most common way to install software), an
SPDN is queried so that the software package management system installs
necessary dependencies of that software package. Unfortunately, to the best of
our knowledge, SPDNs have not hitherto been studied in the context of network
embeddings (despite that they have been studied in the context of data mining
for other purposes, e.g., \cite{smbbgmpa17,dmc18})---the huge literature of
network embeddings have been largely focusing on social networks, citation
networks, and web networks (e.g., \cite{ylzsc15,gf18}).

In this paper: (1) We create two new directed networks with text-associated and
multi-labeled nodes, the package dependency networks in Debian and Fedora, two
popular GNU/Linux distributions. (2) We propose \textit{parent-child
text-associated DeepWalk-1/-2} (PCTADW-1/PCTADW-2), two algorithms for learning
network embeddings for directed networks with text-associated nodes, and we
demonstrate that PCTADW-1 and PCTADW-2 outperformed other algorithms in terms of
effectiveness. (3) The systematic presence of analogies has been long observed
in word embeddings and has played an essential role in human understanding of
word embeddings and algorithmically understanding words (e.g.,~\cite{msccd13}).
Unfortunately, such analogies have not been systematically observed in network
and document embeddings. For the first time, we observe the systematic presence
of analogies in network embeddings. (4) We finally demonstrate that a network embedding can
be novelly used for better understanding \textit{software attributes} (SAs),
such as how software are
developed, software user interface, etc.

\section{Background}

\subsection{Software Package Dependency Network} A \textit{software package dependency
network} (SPDN) characterizes the dependency relationship between software
packages in a software package management system. Simply speaking, a package
\(A\) depends on a package \(B\) iff installing \(A\) requires installing \(B\)
first. They are usually directed networks in which each node represents a
software package and each directed edge characterizes a dependency relation. For
example, in a SPDN that describes Python packages (such as those in PyPI), there
would be an edge that connects the node representing ``tensorflow'' to the node
representing ``numpy'' to represent the fact that the software package
``tensorflow'' depends on ``numpy.'' In addition, in modern software package
management systems, each package is usually associated with text description.

\subsection{Learning Network Embeddings on Networks with Text-Associated Nodes}
As mentioned in the introduction, the majority of current network embedding
learning algorithms focus on the networks alone without taking into account data
associated with nodes, especially text data. TADW is a recently developed
algorithm that learns network embeddings on networks with text-associated nodes.
It learns a network embedding by factorizing the matrix \[M=(A+A^2)/2,\] where
\(A\) is the \(|V|\times |V|\) probabilistic transition matrix characterizing
the transitions of states of a randomly walking agent on the network, i.e.,
\begin{equation}
    (A)_{uv}=\begin{cases}1/\deg(u)&\quad \text{if }(u,v)\in E\\0&\quad
\text{otherwise},\end{cases}
\end{equation}
where \(\deg(u)\) is the degree of \(u\). To
incorporate text associated with nodes, TADW multiplies an additional matrix
that represents text features during the matrix factorization. In other words,
it computes
\begin{equation}
    \min_{W,H}||M-W^{\top}HT||^2_F+\frac{\lambda}{2}\left(||W||^2_F+||H||^2_F\right).
\end{equation}
Here, \(T\) is a matrix consisting of the text feature vector of each node, and
\(W\) and \(H\) are the to-be-learned matrices that consist of vector
representations of each node. From the equation above, it is easy to see that
the space complexity of TADW scales quadratically with respect to the number of
nodes even if the network is sparse and therefore bears a scalability issue.

\section{Embeddings of Directed Networks}

For a given directed network \(G=\langle V, E \rangle\), we represent each
\(v\in V\) using two real number-valued vectors \(\boldsymbol v_c\) and
\(\boldsymbol v_p\). \(\boldsymbol v_c\) and \(\boldsymbol v_p\) encode \(v\)
from two different perspectives---\(v\) as a child (from its incoming edges) and
parent (from its outgoing edges)---and are referred to as the \textit{child
vector representation} and \textit{parent vector representation} of \(v\),
respectively. (\(u\) is a parent (node) of \(v\) and \(v\) is a child (node) of
\(u\) iff \((u, v)\in E\).) We modify the Skip-gram model~\cite{pas14} as to
minimize a negative log probability, i.e., as to compute
\begin{equation}
    \label{eq:goal}
    \begin{split}
        \argmin_{\substack{\boldsymbol v_c,\boldsymbol v_c^{\dagger},
            \boldsymbol v_p, \boldsymbol v_p^{\dagger}: v\in V\\\boldsymbol u_c,
            \boldsymbol u_c^{\dagger}, \boldsymbol u_p,
            \boldsymbol u_p^{\dagger}: u\in V}}L_g(G)=-\sum_{\substack{u,v\in
            V:\\u\neq v}} [&w^c_{u,v}\log p_c(v~|~u)+\\&w^p_{v,u}\log p_p(u~|~v)],
    \end{split}
\end{equation}
where \(w^c_{u, v}\) and \(w^p_{v, u}\) are two non-negative
weighting hyperparameters and have cutoffs (i.e., \(w^c_{u, v}=0\) and
\(w^p_{v,u}=0\) if \(\dist(u,v)>s\), where \(s\) is another positive
hyperparameter and \(\dist(u,v)\) is the distance from \(u\) to \(v\) in the
directed network), and
\begin{align}
  p_c(v~|~u)&=\frac{\exp \left(\boldsymbol v^{\dagger\top}_c\boldsymbol u_p
              \right)}{\sum_{v'\in V}\exp\left( \boldsymbol
              v'^{\dagger\top}_c\boldsymbol u_p\right)}\label{eq:factor-child}\\
  p_p(u~|~v)&=\frac{\exp\left(\boldsymbol u_p^{\dagger\top}\boldsymbol
              v_c\right)}{\sum_{u'\in V}\exp\left(\boldsymbol
              u'^{\dagger\top}_p\boldsymbol v_c\right)}.
  \label{eq:factor-parent}
\end{align}
Here, vectors with and without superscript \(\dagger\) are analogous
to ``output'' and ``input'' vectors in \cite[Equation~(2)]{msccd13},
respectively. \Cref{eq:factor-child,eq:factor-parent} have similar forms to
\cite[Equation~(4)]{pas14} and \cite[Equation~(2)]{msccd13}. Unlike
\cite[Equation~(4)]{pas14} and \cite[Equation~(2)]{msccd13}, which train vectors
to accurately predict surrounding nodes from given nodes without incorporating
directions, we take directions into consideration: \(p_c(v~|~u)\) considers
\(v\) as an \(s\)-child of \(u\) and \(p_p(u~|~v)\) considers \(u\) as an
\(s\)-parent of \(v\) (\(u\) is an \textit{\(s\)-parent (node)} of \(v\) and
\(v\) is an \textit{\(s\)-child (node)} of \(u\) iff \(\dist(u,v)\le s\)). The
model is also flexible enough to allow explicit weighting of the relationship
between two nodes based on various factors such as their distances and the
structure of the network, while \cite{msccd13} does not consider the effects of
distances between two words (as long as they are close within a given window
size), and \cite{pas14} only implicitly incorporated these factors in their
sampled random sequences but does not explicitly discuss or derive their effects
in their optimization goal.

\section{Incorporating Text Associated with Nodes}

While \cref{eq:goal} learns an embedding of a directed network, it does not
incorporate text associated with nodes. To incorporate them, we alter the
optimization goal to
\begin{equation}
    \label{eq:final-goal}
    \argmin_{\substack{\boldsymbol v_c,\boldsymbol v_c^{\dagger}, \boldsymbol
        v_p, \boldsymbol v_p^{\dagger}: v\in V\\\boldsymbol u_c,\boldsymbol
        u_c^{\dagger}, \boldsymbol u_p, \boldsymbol u_p^{\dagger}: u\in V}} L(G, D)=L_g(G) + L_d(D),
\end{equation}
where \(D=\{d_{v}~|~v\in V\}\) is a set of documents and $d_v$ is the document
associated with node $v$. We then have the objective function
\begin{equation}
    L_d(D)=-\sum_{v\in V} \sum_{\omega \in d_v} w_v
    \log\left(\frac{\exp(\boldsymbol \omega^{\top} \boldsymbol v)}{\sum_{\omega'
          \in \Omega}\exp(\boldsymbol \omega'^{\top} \boldsymbol v)}\right),
\end{equation}
where \(\Omega\) is a vocabulary, $w_v$ is a weighting
hyperparameter associated with node $v$, $\omega$ is a word from the document associated with $v$, \(\boldsymbol \omega\) is the representation of $\omega$,
and \(\boldsymbol v\) is a concatenation of \(\boldsymbol v_c\)
and \(\boldsymbol v_p\). This is similar to PV-DBOW~\cite{lm14}: We minimize the
error of predicting a word in a document given the node with which it is
associated.

\section{Architectures of our Neural Networks}

\begin{figure}[!t]
\centering
  \begin{subfigure}[t]{\linewidth} \centering
\includegraphics[width=\linewidth]{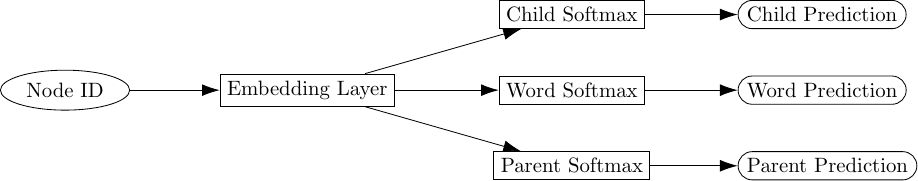}
    \caption{PCTADW-1}\label{fig:architecture-1}
  \end{subfigure} \hfill
  \begin{subfigure}[t]{\linewidth} \centering
\includegraphics[width=\linewidth]{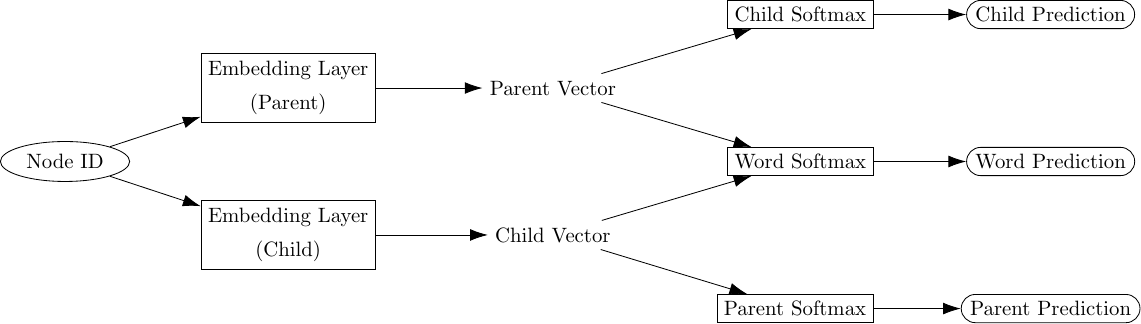}
    \caption{PCTADW-2}\label{fig:architecture-2}
  \end{subfigure}
  \caption{Architecture of PCTADW-1 and PCTADW-2\@. Elliptical units represent
input of data. Units with no border represent intermediate variables.
Rectangular units represent activation units. Rounded rectangular units
represent predictions.}\label{fig:architecture}
\end{figure}

The architecture of our NNs, PCTADW-1 and PCTADW-2, are as follows.
PCTADW-1 approximates \cref{eq:final-goal} with an additional constraint
\(\forall v\in V: \boldsymbol v_c = \boldsymbol v_p\) imposed. PCTADW-1
takes the vector representation of a node as its input. It has three
softmax units that take the input vector directly as input. The first
softmax unit, which we refer to as the \textit{word softmax unit},
predicts a randomly sampled word in the node's associated text. The word
softmax unit learns text associated with nodes. The second/third softmax
unit, which we refer to as the \textit{parent/child node softmax unit},
predicts a randomly sampled \(s\)-parent/child node. The parent/child
node softmax unit learns the structure of the network.
\Cref{fig:architecture-1} illustrates the architecture of PCTADW-1.

PCTADW-2 takes the child and parent vector representations of a node as input.
It approximates \cref{eq:final-goal}. It has three softmax units. The first
softmax unit, the word softmax unit, takes both the two vectors as input to a
softmax unit that predicts a randomly sampled word in the node's associated
text. The second/third softmax unit, the parent/child node softmax unit, takes
the node's child/parent vector representation as its input and predicts a
randomly sampled \(s\)-parent/child node. \Cref{fig:architecture-2} illustrates
the architecture of PCTADW-2.

In both PCTADW-1 and PCTADW-2 to handle nodes with no associated
texts, or parent/child nodes, we simply do not backpropagate from their
respective output nodes.

\section{Sampling of Training Data Points}

In each epoch, we iterate over each node \(v\) once. We feed \(v\) into
the NN for \(t_v\) times, where \(t_v\) is a hyperparameter. We set
\[t_v = \min\{ \max\{ n_p(v), n_c(v) \}, m\}, \]
where \(n_p(v)\) and \(n_c(v)\) are the numbers of \(s\)-parent and
\(s\)-child nodes of \(v\), respectively, and \(m\) is a hyperparameter
that is used to prevent \(t_v\) from being too large.

Each time when feeding \(v\) into the NN, we also
sample a word \(\omega\) in the associated document, an \(s\)-child \(u\) of \(v\), and an
\(s\)-parent \(u'\) of \(v\). \(\omega\) is uniformly randomly sampled in
\(d_v\). \(u\) is sampled by randomly walking from \(v\) along
edge directions for \(s\) nodes and then uniformly randomly choosing a \(u_s\)
in the walked path. \(u'\) is sampled similarly except that we walk
inversely along edge directions.

It is easy to see that \[w_v= \frac{t_v}{l(d_v)},\] where \(l(d_v)\) is the
number of words in \(d_v\). We now derive \(w^c_{u,v}\) and \(w^p_{v,u}\) in
\cref{eq:goal} resulted from our random
walk scheme. During random walk, we denote the state
of being at node \(\nu\) using \(e_{\nu}\), a vector with its
\(\nu^{\text{th}}\) element being 1 and all other elements 0. Let \(A\)
be the transit matrix of the random walk that follows edge
directions, i.e.,
\begin{equation}
  {(A)}_{\mu\nu}=\begin{cases}
    1/\deg_{\text{out}}(\mu)&\quad\text{if }(\mu, \nu)\in E\\
    0&\quad\text{otherwise}
  \end{cases}
\end{equation}
characterizes the probability of transiting from \(e_{\mu}\) to
\(e_{\nu}\). Here, \(\deg_{\text{out}}(\mu)\) is the out-degree of node
\(\mu\). Therefore, for a sufficiently large number of epochs \(k\), when \(v\) is
input to the NN, the expected number of times that \(u\) is the
to-be-predicted \(s\)-child node is \[\frac{k t_v}{s} \left( e_v \sum_{i=1}^s A^i\right)_u.\] Similarly,
for a sufficiently large number of epochs \(k\), the expected number of times that
\(u'\) is the to-be-predicted \(s\)-parent node is
\[\frac{k t_v}{s} \left(e_v\sum_{i=1}^s B^i\right)_{u'},\] where
\begin{equation}
  {(B)}_{\mu\nu}=\begin{cases}
    1/\deg_{\text{in}}(\mu)&\quad\text{if }(\mu, \nu)\in E\\
    0&\quad\text{otherwise},
  \end{cases}
\end{equation}
where \(\deg_{\text{in}}(\mu)\) is the in-degree of node \(\mu\). Since
\(w^c_{u,v}\) and \(w^p_{v,u}\) are proportional to the numbers of times
that \(u\) is used to predict its \(s\)-child \(v\) and that \(v\) is
used to predict its \(s\)-parent \(u\), assuming an asymptotically large
number of epochs, we have
\begin{align}
  w^c_{u,v}&= \frac{t_u}{s}\left( e_u\sum_{i=1}^s A^i\right)_v&\quad w^p_{v,u}&= \frac{t_v}{s}\left( e_v\sum_{i=1}^s B^i\right)_u.
\end{align}

\section{Experimental Evaluation}

In this section, we report our experimental evaluation of the two proposed network embedding algorithms PCTADW-1 and PCTADW-2 with a
focus on SPDNs via node classification.

\begin{table}[t]
\centering
\caption{Statistics of two SPDNs.}
\label{tab:stats}
\begin{tabular}{cccc} 
\toprule
SPDN & Num. of Nodes & Num. of Edges & Num. of Labels\\
\midrule
\textit{Fedora} & \num{49282} & \num{302057} & \num{7} \\
\textit{Debian} & \num{51259} & \num{235982} & \num{8} \\
\bottomrule
\end{tabular}
\end{table}

\subsection{Datasets}

We created 2 datasets,
\textit{Fedora} and \textit{Debian}, each of which is a SPDN whose
most nodes are associated with text. \textit{Fedora} and \textit{Debian}
describe the dependency relationship of software packages
in the GNU/Linux distributions Fedora 28 and Debian 9.5,
respectively. We chose these two GNU/Linux distributions since they both play essential roles in the deployment of GNU/Linux: Fedora is the foundation of Red Hat Enterprise Linux, a popular GNU/Linux distribution, which, for example, powers Summit and Sierra, two out of the three fastest supercomputers in the world as in June 2018~\cite{redhat}; Debian and its derivation, Ubuntu, were reported as the top two choices of operating systems for hosting web services as in 2016~\cite{w3techs}. In these networks, each node represents a software package
and is associated with a description of it. Each edge represents a
dependency relation.
\Cref{tab:stats} shows the statistics of \textit{Fedora} and \textit{Debian} SPDNs.
We generated \textit{Fedora} using DNF Python
API\footnote{ \url{https://dnf.readthedocs.io/en/latest/api.html}} and \textit{Debian} using
python-apt\footnote{ \url{https://salsa.debian.org/apt-team/python-apt}}. We removed all isolated nodes in both networks. We
also manually removed 46 erroneous cyclic dependencies in
\textit{Debian}\footnote{We have made these data publicly available online \anonymizeurl{\url{https://doi.org/10.5281/zenodo.1410669}} .}.

\subsection{Baselines}

\subsubsection{DeepWalk}\cite{pas14} learns vector representations of nodes
in a network using Skip-gram~\cite{msccd13}, a widely used word vector
representation learning algorithm in computational linguistics. It
treats sequences of nodes generated by random walks in the network as
sentences and applies Skip-gram on them. It demonstrated superior
effectiveness compared with a few previous approaches on some social
networks. While it works well on networks, it does not take additional
information, such as those associated with nodes, into account. We used
the implementation by the original authors of
\cite{pas14}\footnote{\url{https://github.com/phanein/deepwalk}}. In our experiments, we set
the dimension of learned vector representations to 128. We used the default values for its other hyperparameters,
i.e., we set the number of random walks to start at each node to 10
and the length of each walk to 40.

\subsubsection{Doc2Vec}~\cite{lm14} learns vector representations for
documents using an NN that is similar to the ones used in
Skip-Gram~\cite{msccd13}: It has an embedding layer after each input
unit, followed by a (hierarchical) softmax unit that predicts a word.
\cite{lm14} proposes two major variants: PV-DM and PV-DBOW\@. PV-DM
trains an NN that takes a document and a randomly sampled window of
words in this document as input and predicts another word in the sampled
window. PV-DBOW trains an NN that takes a document as input and
predicts a word randomly sampled from it. In our experiments, we used these two methods to
generate a vector representation for the document associated with each
node and use it to represent this node. We used the implementation from the \texttt{gensim} software package~\cite{rs10}. We set the dimension of learned vector representations to 128 and the number of training epochs to 100. We also removed stopwords in all documents before we applied PV-DM and PV-DBOW.

\subsubsection{Simple Concatenation} represents a node using a concatenation
of its vector representations learned by DeepWalk and PV-DM or PV-DBOW\@. We used the same hyperparameters for DeepWalk, PV-DM, and PV-DBOW as before except that we set the dimension of the learned vector representations to 64 so that the dimension of the concatenated vectors is 128.

\subsubsection{TADW}~\cite{ylzsc15} learns vector representations of nodes
in a network in which each node is associated with rich text (a document).
We used the vectors learned by PV-DM and PV-DBOW as the feature vectors of documents. For each network, we applied TADW twice with the
dimension of the learned vectors set to 128 (same as the setting of
PCTADW) and 500 (same as in \cite{ylzsc15}). We used the implementation by the OpenNE library\footnote{\url{ https://github.com/thunlp/OpenNE}}.

\subsubsection{Paper2vec}~\cite{sv17} learns vector representations of nodes
in an undirected network \(G\) in which each node is associated with
rich text. Paper2vec had its name because this work was conducted in the
context of research paper citation networks and was also tailored for
them. (a) First, it applies doc2vec on each node's associated text to
train an interim vector representation of the node. (b) Then, it
modifies \(G\) to \(G'\) by adding an edge between each node \(v\) and
its \(\kappa\) nearest neighbors in the interim vector space. (c)
Finally, it applies DeepWalk on \(G'\), in which all vector
representations are initialized to their corresponding interim ones. We
set \(\kappa=4\) because, in \cite{sv17}, this hyperparameter led to the
best effectiveness for networks of ``medium'' sizes, which are similar
to those of \textit{Fedora} and \textit{Debian}. We set the dimension of
learned vector representations to 128. We set the hyperparameters of
DeepWalk to the same as in our DeepWalk baseline, which are also the
same as in \cite{sv17}.

\subsection{Settings of PCTADW-1 and PCTADW-2}

We set the dimension of learned vectors to 128. We set \(m=5\) and
\(s=2\). We used Adam~\cite{kb15} to train our NNs and set the learning
rate, \(\beta_1\), and \(\beta_2\) to 0.001, 0.9, and 0.999. We set
the number of training epochs to 100. We note that, although we used
the same number of epochs as PV-DM and PV-DBOW, our number of training
data points in each epoch is much smaller than those in PV-DM and
PV-DBOW\@: PV-DM and PV-DBOW train their NNs to predict every single
word at least once in each document in each epoch, while PCTADW-1 and
PCTADW-2 only train the NNs to predict a single randomly sampled word
from each document and a single \(s\)-parent and \(s\)-child node of
each node.\footnote{We have made these experimental code available online at
  \anonymizeurl{\url{https://github.com/shudan/PCTADW}} .}

\subsection{Node Classification}

\begin{table*}[t]
\centering
\caption{Experimental results of node classification on \textit{Fedora} and
  \textit{Debian}. DeepWalk+PV-DBOW/DeepWalk+PV-DM refers to simple
  concatenation of DeepWalk learned 64-dimensional vectors and PV-DBOW/PV-DM
  learned 64-dimensional vectors, respectively. The results of TADW are not
  shown due to its onerous requirement of RAM\@. The numbers in the cells are
  micro averages of F1 scores of all labels. The higher the number is, the
  better the corresponding classification result is. In each column, the best
  results and results that are within only 0.002 difference therefrom are
  bold.}\label{tab:node-classification}

\begin{subtable}[t]{\linewidth}
\centering
\caption{\textit{Fedora}}
\begin{tabular}{lcrrrrrr}
\toprule
Nodes Used for Training &\phantom{a}&   5\% & 10\% &     20\% &     25\% &     33\% &     50\% \\
\midrule
PCTADW-1         &&  \textbf{0.918} &  \textbf{0.923} &  \textbf{0.928} &  0.929 &  0.929 &  0.931 \\
PCTADW-2         &&  \textbf{0.917} &  \textbf{0.924} &  \textbf{0.930} &  \textbf{0.932} &  \textbf{0.933} &  \textbf{0.934} \\
DeepWalk         &&  0.856 &  0.860 &  0.864 &  0.868 &  0.868 &  0.868 \\
PV-DBOW          &&  0.647 &  0.676 &  0.695 &  0.702 &  0.708 &  0.713 \\
PV-DM            &&  0.520 &  0.556 &  0.582 &  0.585 &  0.592 &  0.597 \\
DeepWalk+PV-DBOW &&  0.908 &  0.916 &  0.923 &  0.925 &  0.927 &  0.927 \\
DeepWalk+PV-DM   &&  0.870 &  0.888 &  0.899 &  0.900 &  0.901 &  0.904 \\
Paper2vec   &&  0.426 &  0.480 &  0.491 &  0.499 &  0.502 &  0.508 \\
\bottomrule
\end{tabular}
\end{subtable}
\hfill
\begin{subtable}[t]{\linewidth}
\centering
\caption{\textit{Debian}}
\begin{tabular}{lcrrrrrr}
\toprule
Nodes Used for Training &\phantom{a}&   5\% & 10\% &     20\% &     25\% &     33\% &     50\% \\
\midrule
PCTADW-1         &&  \textbf{0.911} &  \textbf{0.917} &  \textbf{0.920} &  \textbf{0.921} &  \textbf{0.923} &  0.923 \\
PCTADW-2         &&  \textbf{0.910} &  \textbf{0.918} &  \textbf{0.922} &  \textbf{0.923} &  \textbf{0.924} &  \textbf{0.925} \\
DeepWalk         &&  0.869 &  0.876 &  0.880 &  0.882 &  0.884 &  0.887 \\
PV-DBOW          &&  0.614 &  0.659 &  0.693 &  0.699 &  0.708 &  0.716 \\
PV-DM            &&  0.432 &  0.480 &  0.510 &  0.515 &  0.522 &  0.525 \\
DeepWalk+PV-DBOW &&  0.902 &  0.913 &  \textbf{0.920} &  \textbf{0.922} &  \textbf{0.924} &  \textbf{0.926} \\
DeepWalk+PV-DM   &&  0.859 &  0.876 &  0.886 &  0.888 &  0.892 &  0.894 \\
Paper2vec   &&  0.461 &  0.494 &  0.516 &  0.524 &  0.531 &  0.541 \\
\bottomrule
\end{tabular}
\end{subtable}

\end{table*}

A common method to evaluate the quality of network embeddings is via linear
classification of nodes. We chose \textit{one-vs-rest logistic regression} as
the linear classifier for evaluation. We applied it to the learned vector
representation and labels of each node. We ran multiple rounds of \(k\)-fold
cross validation with different \(k\)'s to evaluate our algorithms on different
percentages of labeled nodes available for training in the network. Unlike
regular \(k\)-fold cross validation, we used a ``reversed'' version of it:
Instead of using \(k-1\) splits of data for training and 1 split for testing, we
used 1 split for training and \(k-1\) splits for testing. We did this because,
in real-world applications of network embeddings, labeled nodes that are
available for training are often the minority. We ran the logistic regression
algorithm for 100 epochs on each learned embedding on each dataset in each round
of cross validation.

We report our experimental results in \cref{tab:node-classification}. We do not
report the results of TADW since we encountered its aforementioned scalability
issue, which has also been reported elsewhere~\cite{lhcyl17}. From the results,
PCTADW-1 and PCTADW-2 predominantly won on \textit{Fedora}. On \textit{Debian},
the most competitive three algorithms are PCTADW-2, PCTADW-1 and
DeepWalk+PV-DBOW\@. PCTADW-2 virtually won on all percentages of nodes used for
training, and PCTADW-1 also won on all percentages of nodes used for training
except for 50\%, in which case it is also very close to the winner.
DeepWalk+PV-DBOW won on all percentages of nodes used for training except for
5\% and 10\%. Overall, PCTADW-1 and PCTADW-2 were relatively more advantageous
than other algorithms when the percentage of nodes used for training is small.

\section{Analogies in Network Embeddings}

\begin{figure*}[!t]
\centering
  \begin{subfigure}{\columnwidth} \centering
\includegraphics[width=\linewidth]{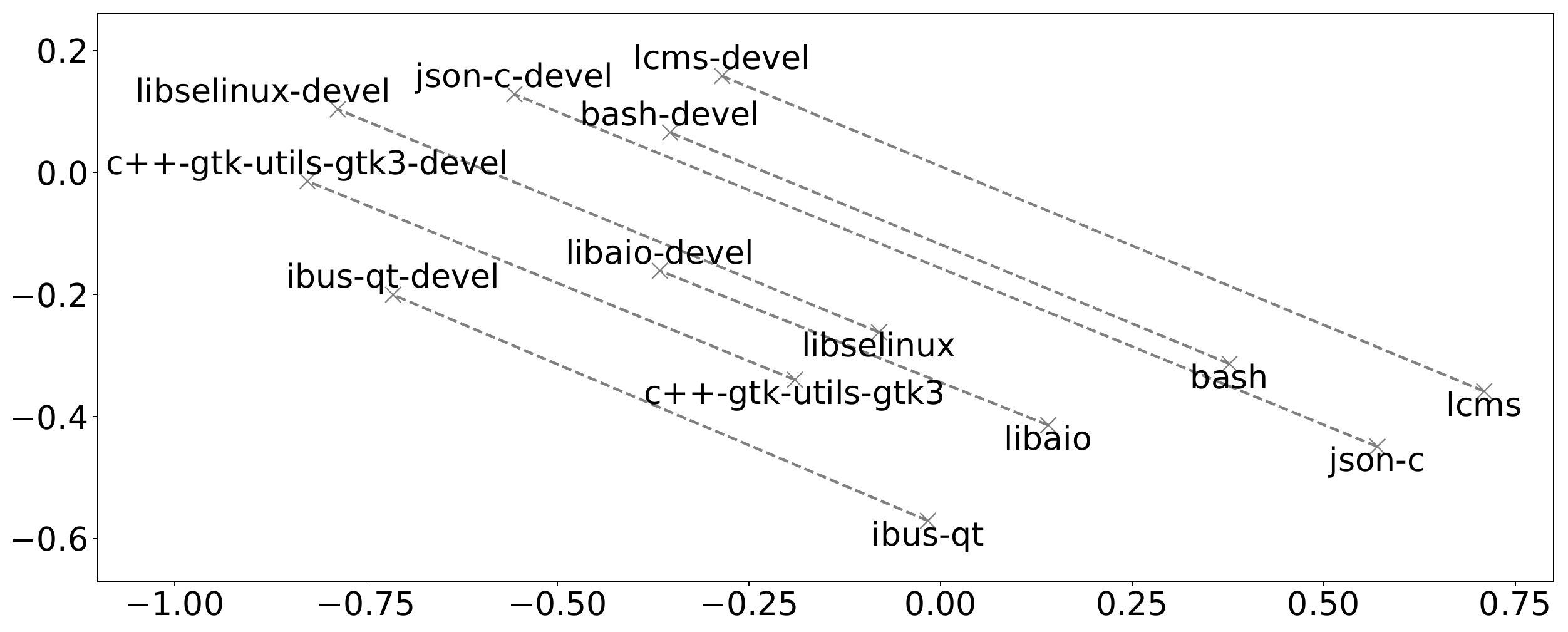}
  \caption{\textit{Fedora}: ``foobar-devel'' vs ``foobar''}
  \end{subfigure} \hfill
  \begin{subfigure}{\columnwidth} \centering
\includegraphics[width=\linewidth]{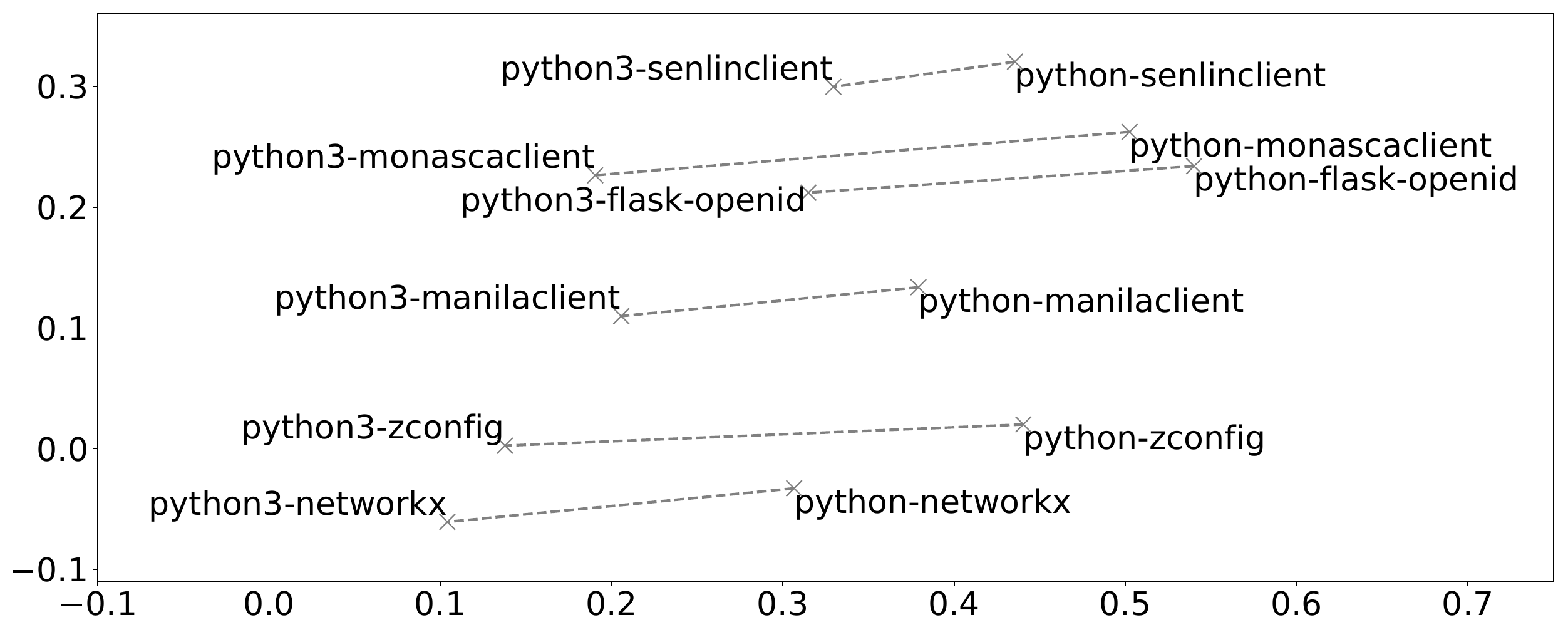}
  \caption{\textit{Debian}: ``python-foobar'' vs ``python3-foobar''}
  \end{subfigure}
  \caption{Illustration of analogies demonstrated in the network embeddings
learned using PCTADW-2. Each figure shows a two-dimensional PCA projection of
the learned 128-dimensional network embeddings. (a) shows the analogies between
development software packages (with their package names ending with ``-devel'')
and non-development software packages (consisting of executables, libraries,
etc.). (b) shows the analogies between Python 2 software packages (with their
package names starting with ``python-'') and their corresponding Python 3
software packages (with their package names starting with ``python3-'') in
Debian.}\label{fig:analogy}
\end{figure*}

\begin{figure}[!t]
    \centering
    \hspace*{\fill}
    \begin{subfigure}[t]{\columnwidth}
    \includegraphics[width=0.9\linewidth]{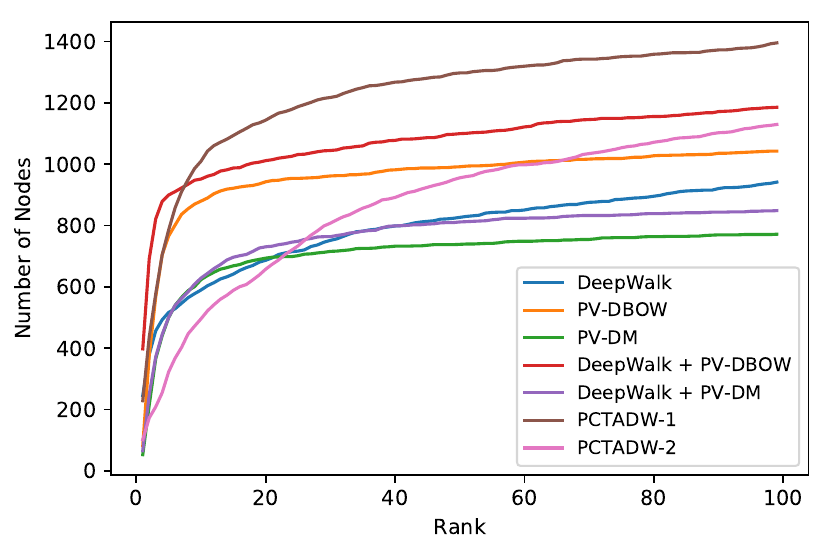}
    \caption{``python-foobar'' vs ``ruby-foobar'' in \textit{Debian}}
    \end{subfigure}
    \hfill
    \begin{subfigure}[t]{\columnwidth}
    \includegraphics[width=0.9\linewidth]{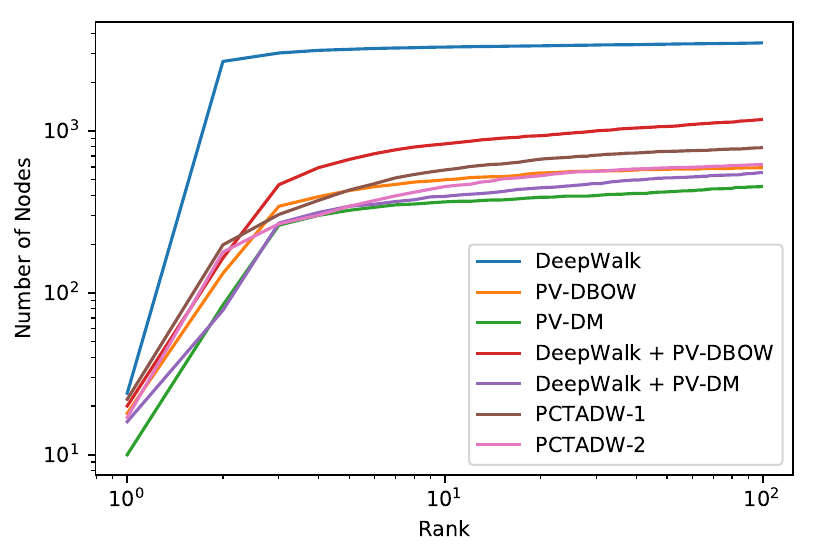}
    \caption{``foobar-devel'' vs ``foobar'' with respect to ``bash-devel'' vs ``bash'' in \textit{Fedora}}
    \end{subfigure}
    \hfill
    \begin{subfigure}[t]{\columnwidth}
    \includegraphics[width=0.9\linewidth]{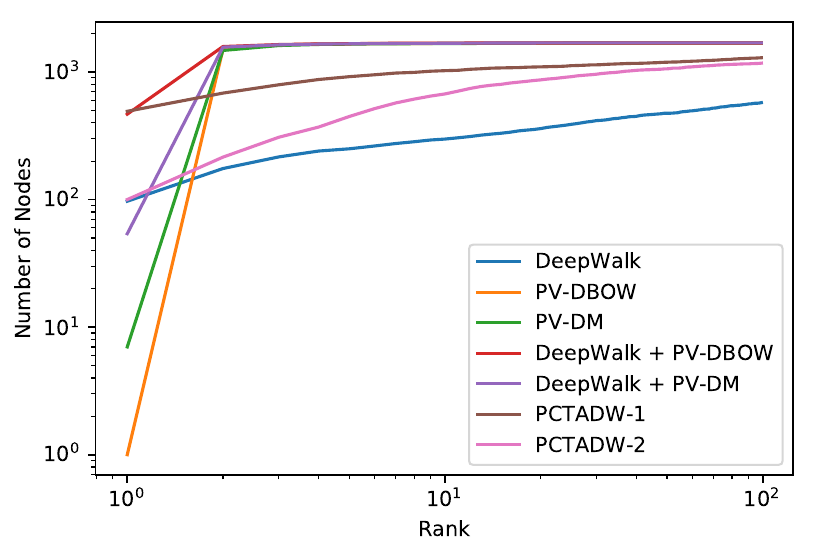}
    \caption{``python-foobar'' vs ``python3-foobar'' with respect to ``python-scipy'' vs ``python3-scipy'' in \textit{Debian}}
    \end{subfigure}
    \hspace*{\fill}
    \caption{Results of our analogy tests. Each plot demonstrates one analogy test. The x-axes represent the rank \(r_b\) and the y-axes represent the number of nodes whose vector representations were ranked below the corresponding rank. Both the x-axes and y-axes in (b) and (c) are in logarithm scales.}
    \label{fig:analogy-tests}
\end{figure}

Analogies in word embeddings have demonstrated their power for facilitating both
algorithmic and human understanding of word embeddings~\cite{msccd13,psm14}.
They have also been used as an empirical instrument for evaluating the quality
of word embeddings~\cite{msccd13}. In a word embedding, an analogy test asks a
question of the form ``WordA1 is to WordA2 what WordB1 is to
\rule{1cm}{0.15mm}.'' \cite{msccd13} shows two examples: `` `Germany' is to
`Berlin' what `France' is to `Paris';'' `` `quick' is to `quickly' what `slow'
to `slowly'.'' Mathematically speaking, these mean that \(\vec
v_{\text{Paris}}\) and \(\vec v_{\text{slowly}}\) are the vectors in the word
embedding that are closest (or very close) to \(\vec v_{\text{Berlin}} -\vec
v_{\text{Germany}} +\vec v_{\text{France}}\) and \(\vec v_{\text{quickly}} -
\vec v_{\text{quick}} + \vec v_{\text{slow}}\), respectively, where \(\vec
v_{\text{word}}\) is the vector reprensentation of ``word.'' Unfortunately, to
the best of our knowledge, similar analogies have not hitherto been discovered
as systematic phenomena in network embeddings or document embeddings. In this
subsection, for the first time, we present the systematic presence of analogies
in network (and document) embeddings in SPDNs.

Software packages often have analogical semantics associated with themselves.
For example, many Python software have both Python 2 and Python 3 versions and
they are usually packaged as two different software packages. Therefore,
semantically we have the analogy of ``Python 2 version of Software A is to its
Python 3 version what Python 2 version of Software B is to its Python 3
version.'' Similar analogies also exist for software with both Python and Ruby
bindings. For another example, many software, especially those written in C or
C\texttt{++}, has two separate software packages: One contains the executables
and libraries of the software and the other, called a development software
package, contains files, such as header files, that are necessary for developing
other C/C\texttt{++} software that uses this software as a depended library. The
names of the latter kind of package always end with ``-devel'' in Fedora and
``-dev'' in Debian. \Cref{fig:analogy} illustrates these analogies.

We performed two types of analogy tests as follows. In the first type of analogy test, on each dataset, we chose some kinds of analogy pairs, such as aforementioned ``foobar-devel'' versus ``foobar'' and ``python-foobar'' versus ``python3-foobar.'' Then, for each kind of analogy pair, we considered every two such pairs \((a_1,a_2)\) and \((b_1,b_2)\). We computed \(\vec v_{a_2} - \vec v_{a_1} + \vec v_{b_1}\) and sorted the distances of all vectors in the network embedding to it. We then recorded the rank \(r_{b_2}\) of \(\vec v_{b_2}\) in these sorted distances. We then compared the cumulative histogram of \(r_{b_2}\) for the network embedding produced by each algorithm. The taller the histogram is, the better the corresponding network embedding performed for that analogy test. The second type of analogy test is similar to the first type, except that, instead of considering every two pairs \((a_1,a_2)\) and \((b_1,b_2)\), we only considered all pairs with a specific given pair. For example, for the kind of analogy pair ``python-foobar'' versus ``python3-foobar,'' we only compared the cumulative histogram of \(r_{b_2}\) in case of \((a_1, a_2)\) being (``python-scipy'', ``python3-scipy''). In this example, we refer to this type of analogy test as ``python-foobar'' versus ``python3-foobar'' with respect to ``python-scipy'' versus ``python3-scipy.'' We used the second type of analogy test when it would be too computationally expensive if its corresponding analogy test of the first type were performed.

We performed three analogy tests, (a) ``python-foobar'' versus ``ruby-foobar'' in \textit{Debian}, (b) ``foobar-devel'' versus ``foobar'' with respect to ``bash-devel'' versus ``bash'' in \textit{Fedora}, and (c) ``python-foobar'' versus ``python3-foobar'' with respect to ``python-scipy'' versus ``python3-scipy'' in \textit{Debian}. \Cref{fig:analogy-tests} shows our experimental results. In (a), PCTADW-1 was the winner. In (b), DeepWalk was the winner. In (c), PV-DBOW, PV-DM, PB DeepWalk+PV-DBOW, and DeepWalk+PV-DM won all other three algorithms.

\begin{table}[t]
    \centering
    \caption{Experimental results of predicting nodes corresponding to development software packages in \textit{Fedora} using logistic regression. The first column and all numbers mean the same as in \cref{tab:node-classification}.}
    \label{tab:specific-label-test}
   \begin{tabular}{l|crrrrrr}
    \toprule
\% &\phantom{a}&   PCTADW-1 & PCTADW-2 & DeepWalk \\
\midrule
5\% &&  0.9400 &  0.9376&  0.8354\\
10\% &&  0.9426 &  0.9419 &  0.8325\\
20\% &&  0.9451 &  0.9459 &  0.8294\\
25\% &&  0.9463 & 0.9462 & 0.8293\\
33\% &&  0.9468 & 0.9473 & 0.8289\\
50\% &&  0.9476 & 0.9482 & 0.8277\\
\bottomrule
    \end{tabular}
\end{table}

Our first observation is that, an algorithm won in node classification did not necessarily win in analogy tests. In other words, an algorithm's effectiveness in node classification and analogy tests were not necessarily consistent. For example, although PCTADW-1 and PCTADW-2 outperformed other network embedding learning algorithms, they still lost in analogy tests (b) and (c). In fact, \cref{tab:specific-label-test} briefly shows that both of PCTADW-1 and PCTADW-2 are more effective than DeepWalk in predicting whether a node corresponds to a development package in \textit{Fedora}.

Our second obversation is that, analogy tests were more responsive to what information had been used for training. In \textit{Fedora}, a node representing a development software package ``foobar-devel'' is always directly connected to the node representing ``foobar.'' This strong relationship in the structure of the network could be the reason that led to the best performance of DeepWalk in (b), since DeepWalk produced network embeddings only based on the structure of the network. Similarly, in \textit{Debian}, the description of a package ``python-foobar'' is often almost identical to ``python3-foobar'' with some ``Python'' or ``Python 2'' replaced with ``Python 3.'' This strong similarity in the documents could be the reason that resulted in the winning of all algorithms involving PV-DBOW and PV-DM in (c). While in (a), neither the structure of the network or documents associated with nodes present a strong relationship. In this case, PCTADW-2 won, perhaps because it has the best integration of the structure of the network and documents associated with nodes.

\section{Software Attributes in Network Embeddings}
In this section, we demonstrate how we can use the network embeddings of SPDNs for understanding SAs. We show how these network embeddings
encapsulate information of SAs and how such information can be further used
to understand SAs.

\subsection{Encapsulation of Software Attributes}

\begin{figure}
\scalebox{0.85}{
\begin{tikzpicture}

    \begin{scope}[xslant=-1.4]

        \draw[help lines, fill=gray!30,opacity=0.3] (-0.5, 0) rectangle (3.5, 2);
        \node[circle,fill,inner sep=1pt, label={[align=center]left:  $\vec{r}$ } ](A1) at (1.5, 3){};
        \node[circle,fill,inner sep=1pt, label={[align=center]right:  $\vec{s}$}](B1) at (1.5, -1){};
        \node[cloud, cloud puffs=13, cloud ignores aspect, align=center, draw, minimum width=2cm, minimum height=1.1cm, name path=CA, ball color=gray,
        opacity=0.3](A) at (1.5, 3) {};
        \node[cloud, cloud puffs=13, cloud ignores aspect, align=center, draw, minimum width=2.5cm, minimum height=1.5cm, name path=CB,
        ball color=gray, opacity=0.3] (B) at (1.5, -1) {};

        \coordinate (A2) at (1.5, 3);
        \coordinate (B2) at (1.5, -1);
        \path [name path = direction] (A1) -- (B1);
        \path [name intersections = {of=CA and direction, by={C1}}];
        \path [name intersections = {of=CB and direction, by={C2}}];
        \coordinate (C3) at ($(A1)!0.6!(C1)$);
        \draw[dashed, very thick] (A1) -- (C3);
        \draw[very thick] (C3) -- (C2);
        \draw[->, dashed, very thick] (C2) -- (B1);

        \path [name path = midrow] (A2) -- (B2);
        \path [name path = midrow1] ([xshift=-0.4cm]A2) -- ([xshift=-0.4cm]B2);
        \path [name path = midrow2] ([xshift=0.4cm]A2) -- ([xshift=0.4cm]B2);

        \coordinate (Z) at (-1.2, 1.8);
        \coordinate (Z1) at (1.5, 1.6);
        \path [name path = zsh] (Z) -- (Z1);
        \path [name path = PZ1] ([yshift=-0.2cm]Z) -- ([yshift=-0.2cm]Z1);

        \path [name intersections = {of=PZ1 and midrow, by={Z2}}];
        \path [name intersections = {of=PZ1 and midrow1, by={Z3}}];
        \path [name intersections = {of=zsh and midrow1, by={Z4}}];

        \node[circle,fill,inner sep=0.8pt] at (Z){};
        \draw[help lines] (Z2) -- (Z3) -- (Z4);

        \coordinate (G) at (-0.8, -0.7);
        \coordinate (G1) at (1.5, 1.2);
        \path [name path = git] (G) -- (G1);
        \path [name path = PG1] ([yshift=-0.2cm]G) -- ([yshift=-0.2cm]G1);

        \path [name intersections = {of=PG1 and midrow, by={G2}}];
        \path [name intersections = {of=PG1 and midrow1, by={G3}}];
        \path [name intersections = {of=git and midrow1, by={G4}}];

        \node[circle,fill,inner sep=0.8pt] at (G){};
        \draw[help lines] (G2) -- (G3) -- (G4);

        \coordinate (D) at (6.4, 3);
        \coordinate (D1) at (1.5, 0.2);

        \path [name path = gdb] (D) -- (D1);
        \path [name path = PD1] ([yshift=0.2cm]D) -- ([yshift=0.2cm]D1);

        \path [name intersections = {of=PD1 and midrow, by={D2}}];
        \path [name intersections = {of=PD1 and midrow2, by={D3}}];
        \path [name intersections = {of=gdb and midrow2, by={D4}}];

        \node[circle,fill,inner sep=0.8pt] at (D){};
        \draw[help lines] (D2) -- (D3) -- (D4);

        \path [name path = rectangle] (0, 0) rectangle (2, 2);

        \path [name intersections = {of = rectangle and git, by={I1}}];

        \draw[thick](Z)node[left]{\texttt{zsh}} -- (Z1);

        \draw[thick](G)node[below]{\texttt{git}} -- (I1);

        \draw[thick](D)node[above]{\texttt{gdb}} -- (D1);
        \draw[dashed, thick](I1) -- (G1);

    \end{scope}

\end{tikzpicture}
}
    \caption{Illustration of the encapsulation of the SA \textsc{Regular-User/Developer}. For visualization purpose, we assume that the dimension of the network embedding is 3. Each cloud represents a reference software package collection. The two dots inside the clouds represent the median vectors of all vector representations in the two collections, respectively. The arrow from \(\vec r\) to \(\vec s\) represents the direction that characterizes the SA. The projections of three example software packages \texttt{zsh}, \texttt{git}, and \texttt{gdb} are shown. To help readers visualize, we draw a plane (the gray area) that passes through \((\vec s - \vec r)\).}
    \label{fig:flow}
\end{figure}

\begin{figure}[!t]
    \centering
    \begin{subfigure}[t]{\linewidth}
        \centering
        {\textsc{Regular-User} \(\longleftarrow\) \(\longrightarrow\) \textsc{Developer}}
       \includegraphics[width=0.95\linewidth]{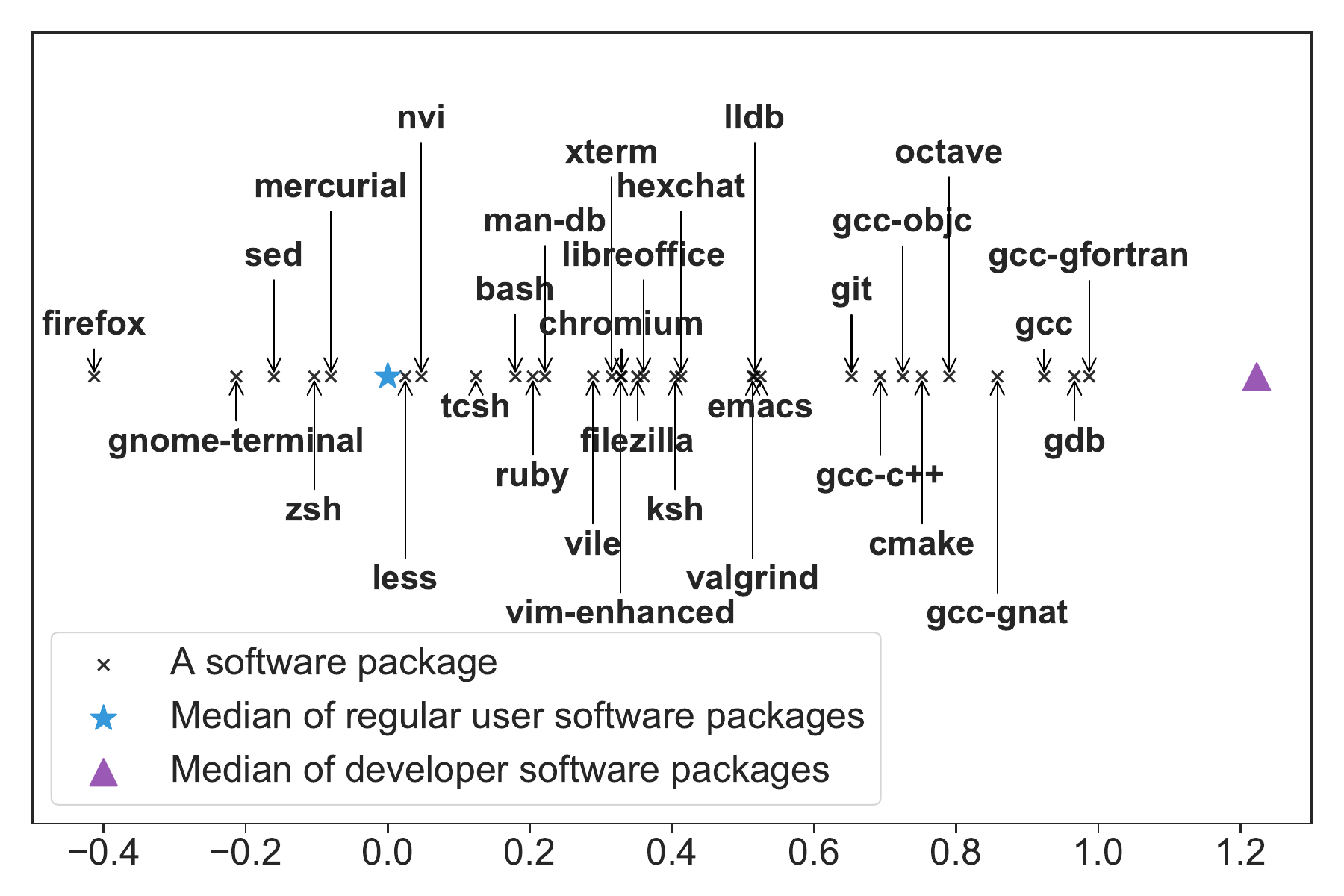}
        \caption{\textit{Fedora}: \textsc{Regular-User/Developer}}
        \label{fig:project_developer}
    \end{subfigure}
    \hfill
    \begin{subfigure}[t]{\linewidth}
        \centering
        {\textsc{Compiled} \(\longleftarrow\) \(\longrightarrow\) \textsc{Interpreted}}
        \includegraphics[width=0.95\linewidth]{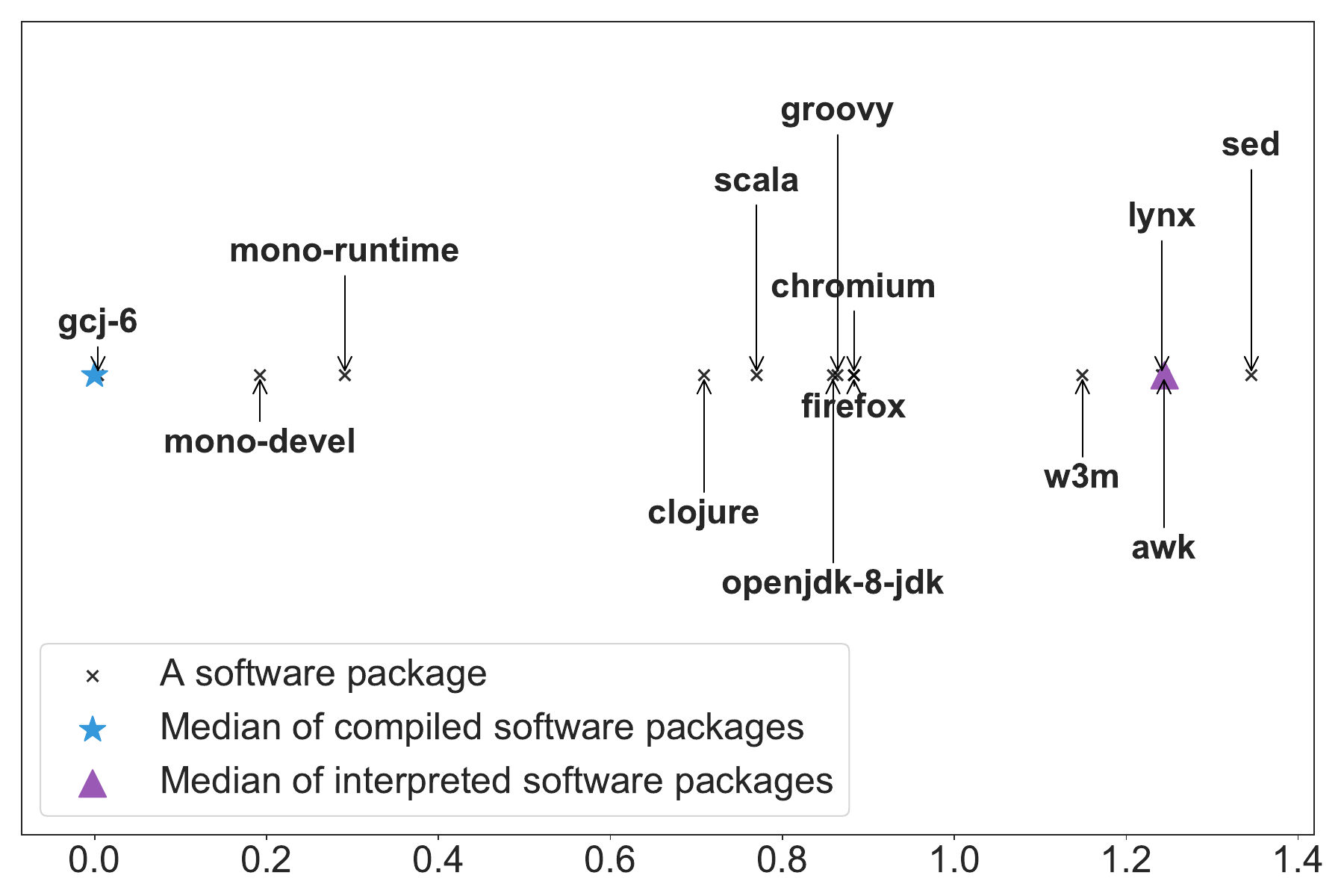}
        \caption{\textit{Debian}: \textsc{Compiled/Interpreted}}
       \label{fig:project_interpret}
    \end{subfigure}
    \caption{Projected positions of software packages for two SAs. Horizontal
      axes represent projected positions. Each cross represents a software
      package, whose name is indicated by an arrow. In each plot, the triangle
      and star represent opposite extremes of the SA.}

    \label{fig:project}
\end{figure}

In this subsection, we show that a direction in the
network embedding space encapsulates information about an SA\@. We do this by
projecting the vector representations of software packages onto individual
directions and analyzing their projected positions on them.
\subsubsection{Identify a direction for a specific SA using reference software
  packages} Given an SA, we first select two \textit{reference} software package
collections, each of which comprises software packages that are quintessences of
one extreme of this SA. We then use the \textit{median vector} of all vector
representations of software packages in each software package collection to
approximate one extreme of this SA. Here, the median vector of a set of vectors
\(\{\vec u_1, \vec u_2, \ldots\}\) is defined as \(\tilde {\vec u}=(\tilde
u^{[1]}, \tilde u^{[2]}, \ldots)\), where each superscript \([\cdot]\) indicates
the index of a component of a vector and \(\tilde u^{[i]}\) is the median of
\(\{u_1^{[i]}, u_2^{[i]}, \ldots\}\) for each $i\in \{1,2,\ldots\}$. We denote
these two median vectors by $\vec{r}$ and $\vec{s}$, respectively. The direction
along \((\vec s-\vec r)\) can be used to characterize the given SA. Here, the
intuition is that the two median vectors are the centroid vectors that best
characterize the two extremes of the SA, and linear combinations of these two
vectors characterize values of the SA that are in between these two extremes.

For example, let us consider the \textsc{Regular-User/Developer} SA, which
characterizes whether a software package is more commonly used by a regular user
or a developer. We select two reference software package collections. One
consists of software packages that are normally used only by developers, such as
\texttt{python-dev}, \texttt{zsh-dev}, and \texttt{ruby2.3-dev}; the other one
consists of software packages that are normally used only by regular users, such
as \texttt{inkscape}, \texttt{libreoffice}, and \texttt{thunderbird}. In this
example, \(\vec r\) would be the median vector of the former collection and
\(\vec s\) would be the median vector of the latter collection. The direction
along \((\vec s - \vec r)\) characterizes the \textsc{Regular-User/Developer}
SA.

\subsubsection{Analyze vector representations of software packages projected
  onto the direction along \((\vec s - \vec r)\)} We first project the vector
representation of each software package onto the direction along \((\vec s -
\vec r)\). The projection results in a projected position
$(\vec{v_p}-\vec{r})\cdot\frac{\vec s - \vec r}{\vert \vec s - \vec r \vert}$
for the vector representation \(\vec v_p\) of each software package \(p\). These
projections extract information about the given SA and largely eliminate the
influence of other SAs over the vector representations. We can then analyze the
given SA through the projected positions: A small projected position indicates
closeness to one extreme of the SA and a larger one indicates the other.

\Cref{fig:flow} illustrates the above steps.
As a case study, we performed these steps on two SAs: \textsc{Regular-user/Developer} and \textsc{Compiled/Interpreted} (which characterizes whether a software package is mainly written in a compiled or interpreted programming language).
\Cref{fig:project} shows the resulting projected positions.
We discuss the result of each SA as follows.

\begin{figure*}[t!]
    \begin{subfigure}[t]{0.47\linewidth}
        \centering
       \includegraphics[width=0.85\linewidth]{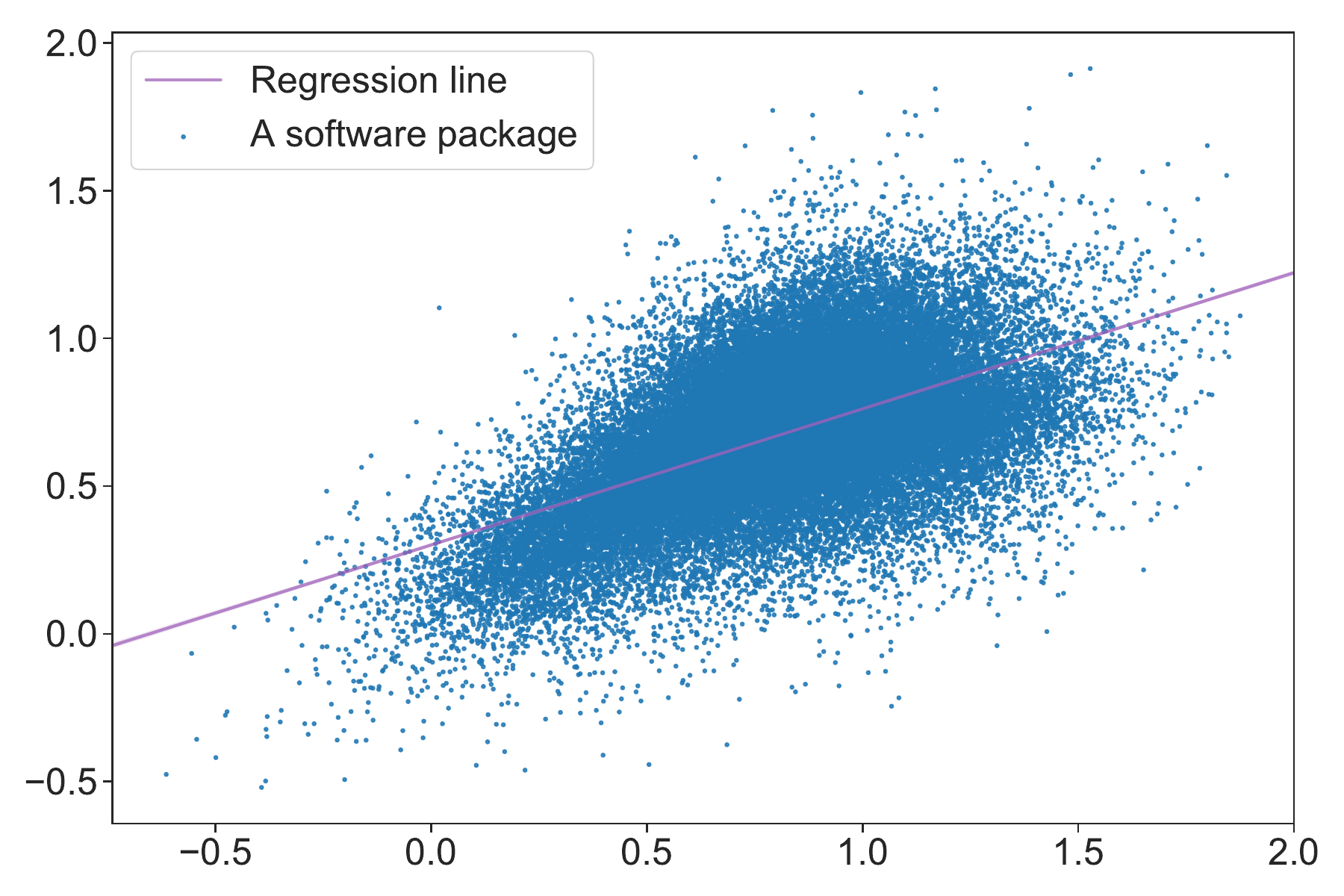}
        \caption{ }
        \label{fig:correlation_regular}
    \end{subfigure}
    \begin{subfigure}[t]{0.47\linewidth}
        \centering
       \includegraphics[width=0.85\linewidth]{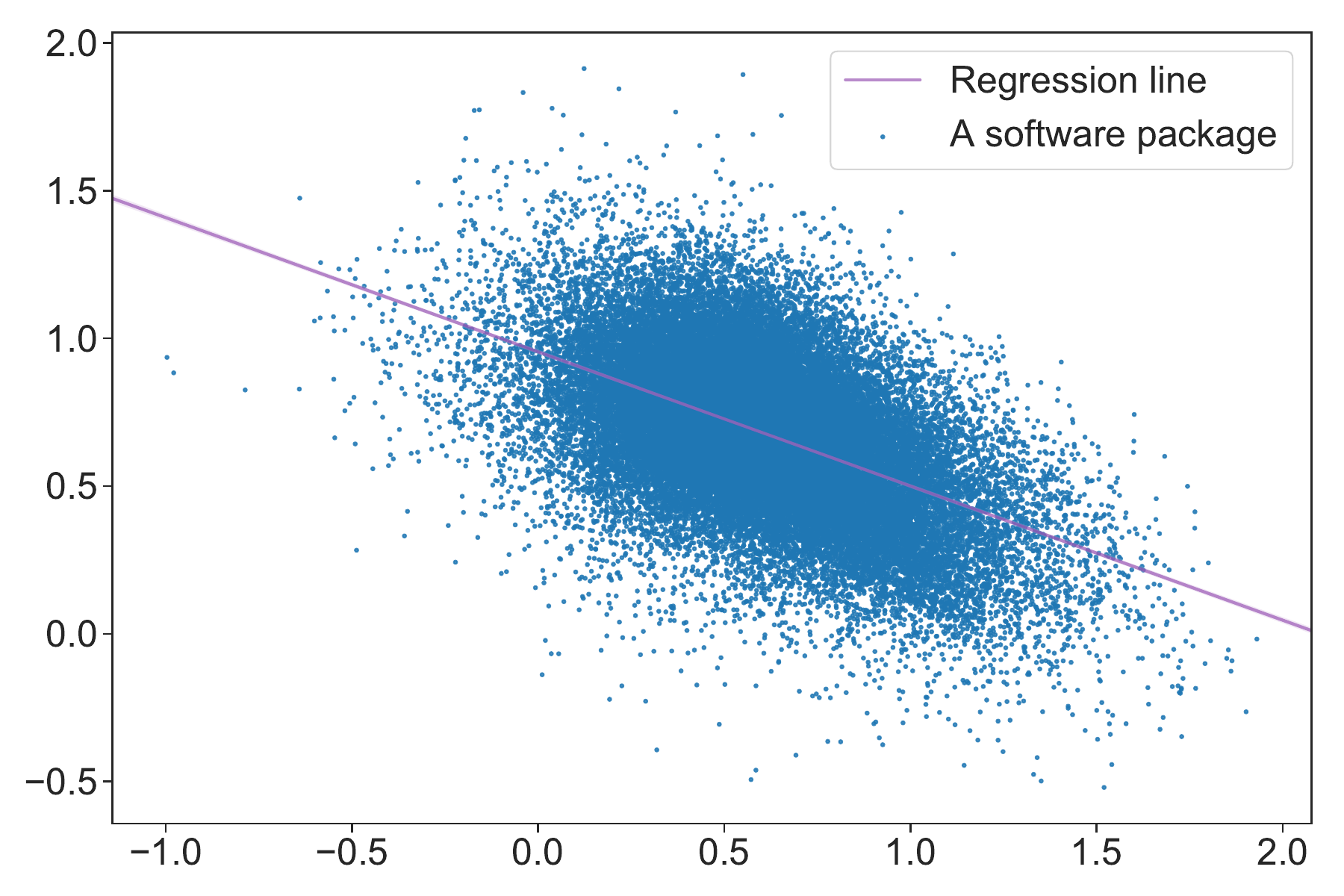}
        \caption{}
        \label{fig:correlation_tui}
    \end{subfigure}
    \caption{Illustration of how the knowledge of the correlation
      between SAs is encapsulated in network embeddings. Each dot
      represents a software package. The \(x\)-axes and \(y\)-axes
      indicate the projected positions onto the direction of two SAs.
      The slopes of the regression lines indicate the correlation
      between the two SAs. A positive slope, like in (a), indicates a
      positive correlation between the two SAs, and a negative slope,
      like in (b), indicates a negative correlation. (a): \(x\)-axis: \textsc{Regular-User/Developer};\(y\)-axis: \textsc{Client/Server}. (b): \(x\)-axis: \textsc{TUI/GUI}; \(y\)-axis: \textsc{Client/Server}}
    \label{fig:correlation}
\end{figure*}

\paragraph{\textbf{\textsc{Regular-User/Developer}}}
We used the \textit{Fedora} SPDN. In the reference software package collection of \textsc{Regular-User}, we put quintessential non-development software packages, such as \texttt{thunderbird} and \texttt{inkscape}; in the collection of \textsc{Developer}, we put all software packages whose names end with ``-devel,'' which consist of mostly C/C\texttt{++} header files.

\Cref{fig:project_developer} shows the result. We selected the shown software packages because their \textsc{Regular-User/Developer} SAs are representative in terms of their closeness to the two extremes: Some are supposedly close to \textsc{Regular-User}, some are close to \textsc{Developer}, and some are in between. None of them were included in either reference software package collections.
With a few exceptions (such as \texttt{firefox}), software packages with small projected positions are generally more commonly used by regular users and those with large projected positions are more commonly used by developers. For example, \texttt{gcc-c++}, \texttt{cmake}, and \texttt{gdb} have larger projected positions because they are commonly used in software development; \texttt{zsh}, \texttt{sed}, and \texttt{less}, have smaller projected positions because they usually serve as system utilities; \texttt{emacs} has a larger projected position than that of \texttt{vim-enhanced} because, in addition to editing programming code, \texttt{vim-enhanced} is also commonly used in system administrative tasks while \texttt{emacs} is not.

\paragraph{\textbf{\textsc{Compiled/Interpreted}}}
We used the \textit{Debian} SPDN\@. In the reference software package collection of \textsc{Compiled} and \textsc{Interpreted}, we respectively put quintessential compiler software packages, such as \texttt{go} and \texttt{gcc-c++}, and interpreter software packages, such as \texttt{python} and \texttt{ruby}.

\Cref{fig:project_interpret} shows the result. We selected the shown software packages with a criteria similar to that of \cref{fig:project_developer}. Small projected positions generally indicate closeness to compilers, and large projected positions indicate closeness to interpreters. For example, all software packages left to \texttt{groovy} provide environments for programming languages that commonly compiled to byte code, which are interpreted during runtime, while those on the right directly interpret programming or markup languages.

\subsection{Interactions between Software Attributes}
In this subsection, we demonstrate that network embeddings can be
leveraged to analyze interactions between SAs. For any two SAs, we do
this by statistically measure the relationship between the projected
positions of all software packages onto the two corresponding directions.

For every two SAs, similar to the steps in the previous subsection, we
first identify the directions of SAs and obtain projected positions onto
the two directions for all software packages. We then measure the
correlation between the SAs of these software packages using the Pearson
correlation coefficient. Each plot in \cref{fig:correlation} illustrates
the projected positions for all software packages onto the directions of
two SAs, and how a regression line indicates the correlation between
them.

\Cref{tab:correlation} summarizes the correlation matrix of various SAs,
computed from the projected positions of all software packages onto the
directions of these SAs. With the exception of \textsc{TUI/GUI}, the
signs of the Pearson correlation coefficients are mostly consistent with
human knowledge. For example, \textsc{Regular-User/Developer} and
\textsc{Work/Entertainment} are negatively correlated, which is
consistent with the fact that software packages for work and not
entertainment are more likely to be used by developers than regular
users, because work on a GNU/Linux operating system is very often
associated with software development. The positive correlation between
\textsc{Regular-User/Developer} and \textsc{Client/Server} is consistent
with the fact that regular users usually only use software packages that
run on the client side and developers also use software packages that
run on the server side.

This correlation matrix can also be used to discover unknown and
unobvious relationship between SAs. For example,
\textsc{Regular-User/Developer} and \textsc{Compiled/Interpreted} are
negatively correlated. This is consistent with the fact that 18 out of
the 22 most popular development environments as of Feb, 2019 (as
reported in \cite{stackoverflow}) are primarily written in (partially)
compiled programming languages, such as Java and C/C\texttt{++}.

\begin{table}[t]
   \caption{Pearson correlation coefficients between various SAs. Regular,
     Compiled, TUI, Client, and Work
     represent SAs \textsc{Regular-User/Developer},
     \textsc{Compiled/Interpreted}, \textsc{TUI/GUI}, \textsc{Client/Server},
     and \textsc{Work/Entertainment}, respectively. Each cell shows two Pearson
     correlation coefficients
     between the SAs of the corresponding row and column for \textit{Fedora} and
     \textit{Debian}, respectively. A positive/negative number indicates a
     positive/negative correlation. The larger the absolute value of the number
     is, the stronger the correlation is.}
   \label{tab:correlation}
   \begin{subtable}[t]{\columnwidth}
       \centering
       \caption{\textit{Fedora}}
       \begin{tabular}{ccccccc} 
         \toprule
         & Regular & Compiled & TUI & Client & Work\\
         \midrule
         Regular & 1.00 & -0.28 & 0.06 & 0.26 & -0.15\\
         Compiled & -0.28 & 1.00 & 0.15 & -0.02 & 0.01\\
         TUI & 0.06 & 0.15 & 1.00 & 0.11 & 0.19\\
         Client & 0.26 & -0.02 & 0.11 & 1.00 & 0.09\\
         Work & -0.15 & 0.01 & 0.19 & 0.09 & 1.00\\
         \bottomrule
       \end{tabular}
   \end{subtable}
   \hfill
   \begin{subtable}[t]{\columnwidth}
       \centering
       \caption{\textit{Debian}}
       \begin{tabular}{ccccccc} 
         \toprule
         & Regular & Compiled & TUI & Client & Work\\
         \midrule
         Regular & 1.00 & -0.42 & -0.33 & 0.55 & -0.17\\
         Compiled & -0.42 & 1.00 & -0.21 & 0.10 & 0.03\\
         TUI & -0.33 & -0.21 & 1.00 & -0.52 & 0.09\\
         Client & 0.55 & 0.10 & -0.52 & 1.00 & -0.10\\
         Work & -0.17 & 0.03 & 0.09 & -0.10 & 1.00\\
         \bottomrule
       \end{tabular}
   \end{subtable}
\end{table}

\section{Conclusion and Future Work}

In this paper, we presented PCTADW-1 and PCTADW-2, two algorithms for
learning embeddings of networks with text-associated nodes, including
SPDNs. We extracted two SPDNs from \textit{Fedora} and \textit{Debian}.
We then demonstrated the effectiveness of PCTADW-1 and PCTADW-2 in
comparison with some baselines on these two SPDNs. For the first time,
we discovered and discussed the systematic presence of analogies (that
are similar to those in word embeddings) in network embeddings. We then
demonstrated how we can leverage network embeddings generated from
the SPDNs for understanding SAs. We empirically showed that a direction
in the network embedding vector space encapsulates information about an
SA. We showed that projected positions of software packages on it encode
knowledge about the SA of the software package. We finally demonstrated
that they can also be used to discover and study interactions and
relationship between SAs.

Our work can potentially elicit interesting future research to explore
the use of embedding techniques in the research of SE\@. One
direction is to incorporate other metadata of software packages
during the training procedure. Another direction is to apply our method for understanding SAs to
improve features such as recommendation systems in existing open source
software package management systems. Yet another direction is to extend
our method to other types of datasets in SE, such as timeline graphs in
version control systems.

{
\bibliography{refs}

\begin{thebibliography}{10}
\providecommand{\url}[1]{#1}
\csname url@samestyle\endcsname
\providecommand{\newblock}{\relax}
\providecommand{\bibinfo}[2]{#2}
\providecommand{\BIBentrySTDinterwordspacing}{\spaceskip=0pt\relax}
\providecommand{\BIBentryALTinterwordstretchfactor}{4}
\providecommand{\BIBentryALTinterwordspacing}{\spaceskip=\fontdimen2\font plus
\BIBentryALTinterwordstretchfactor\fontdimen3\font minus
  \fontdimen4\font\relax}
\providecommand{\BIBforeignlanguage}[2]{{%
\expandafter\ifx\csname l@#1\endcsname\relax
\typeout{** WARNING: IEEEtran.bst: No hyphenation pattern has been}%
\typeout{** loaded for the language `#1'. Using the pattern for}%
\typeout{** the default language instead.}%
\else
\language=\csname l@#1\endcsname
\fi
#2}}
\providecommand{\BIBdecl}{\relax}
\BIBdecl

\bibitem{snbgge08}
P.~Sen, G.~Namata, M.~Bilgic, L.~Getoor, B.~Galligher, and T.~Eliassi-Rad,
  ``Collective classification in network data,'' \emph{AI Magazine}, vol.~29,
  no.~3, pp. 93--106, 2008.

\bibitem{adwf15}
B.~Alipanahi, A.~Delong, M.~T. Weirauch, and B.~J. Frey, ``Predicting the
  sequence specificities of {DNA}- and {RNA}-binding proteins by deep
  learning,'' \emph{Nature Biotechnology}, vol.~33, pp. 831--838, 2015.

\bibitem{xskk18}
H.~Xu, K.~Sun, S.~Koenig, and T.~K.~S. Kumar, ``A warning propagation-based
  linear-time-and-space algorithm for the minimum vertex cover problem on giant
  graphs,'' in \emph{the International Conference on the Integration of
  Constraint Programming, Artificial Intelligence, and Operations Research},
  2018, pp. 567--584.

\bibitem{gf18}
P.~Goyal and E.~Ferrara, ``Graph embedding techniques, applications, and
  performance: A survey,'' \emph{Knowledge-Based Systems}, vol. 151, pp.
  78--94, 2018.

\bibitem{pas14}
B.~Perozzi, R.~Al-Rfou, and S.~Skiena, ``{DeepWalk}: {O}nline learning of
  social representations,'' in \emph{the ACM SIGKDD International Conference on
  Knowledge Discovery and Data Mining}, 2014, pp. 701--710.

\bibitem{apaa2018}
S.~Abu-El-Haija, B.~Perozzi, R.~Al-Rfou, and A.~Alemi, ``Watch your step:
  Learning node embeddings via graph attention,'' in \emph{Proceedings of the
  32nd International Conference on Neural Information Processing Systems},
  2018, p. 9198–9208.

\bibitem{hhhs2020}
D.~Huang, Z.~He, Y.~Huang, K.~Sun, S.~Abu-El-Haija, B.~Perozzi, K.~Lerman,
  F.~Morstatter, and A.~Galstyan, ``Graph embedding with personalized context
  distribution,'' in \emph{Companion Proceedings of the Web Conference 2020},
  2020, p. 655–661.

\bibitem{tqwzym15}
J.~Tang, M.~Qu, M.~Wang, M.~Zhang, J.~Yan, and Q.~Mei, ``{LINE}: {Large}-scale
  information network embedding,'' in \emph{the International Conference on
  World Wide Web}, 2015, pp. 1067--1077.

\bibitem{gl16}
A.~Grover and J.~Leskovec, ``node2vec: {Scalable} feature learning for
  networks,'' in \emph{the ACM SIGKDD International Conference on Knowledge
  Discovery and Data Mining}, 2016, pp. 855--864.

\bibitem{clx16}
S.~Cao, W.~Lu, and Q.~Xu, ``Deep neural networks for learning graph
  representations,'' in \emph{the AAAI Conference on Artificial Intelligence},
  2016, pp. 1145--1152.

\bibitem{dcs17}
Y.~Dong, N.~V. Chawla, and A.~Swami, ``Metapath2vec: Scalable representation
  learning for heterogeneous networks,'' in \emph{the ACM SIGKDD International
  Conference on Knowledge Discovery and Data Mining}, 2017, pp. 135--144.

\bibitem{wcwpzy17}
X.~Wang, P.~Cui, J.~Wang, J.~Pei, W.~Zhu, and S.~Yang, ``Community preserving
  network embedding,'' in \emph{the AAAI Conference on Artificial
  Intelligence}, 2017, pp. 203--209.

\bibitem{lhcyl17}
Y.-A. Lai, C.-C. Hsu, W.~H. Chen, M.-Y. Yeh, and S.-D. Lin, ``{PRUNE}:
  {Preserving} proximity and global ranking for network embedding,'' in
  \emph{the Neural Information Processing Systems Conference}, 2017, pp.
  5257--5266.

\bibitem{hyl17}
W.~Hamilton, Z.~Ying, and J.~Leskovec, ``Inductive representation learning on
  large graphs,'' in \emph{the Neural Information Processing Systems
  Conference}, 2017, pp. 1024--1034.

\bibitem{hlh17}
X.~Huang, J.~Li, and X.~Hu, ``Label informed attributed network embedding,'' in
  \emph{the ACM International Conference on Web Search and Data Mining}, 2017,
  pp. 731--739.

\bibitem{wywwwl18}
Z.~Wang, X.~Ye, C.~Wang, Y.~Wu, C.~Wang, and K.~Liang, ``{RSDNE}: {E}xploring
  relaxed similarity and dissimilarity from completely-imbalanced labels for
  network embedding,'' in \emph{the AAAI Conference on Artificial
  Intelligence}, 2018, pp. 475--482.

\bibitem{dai18}
Q.~Dai, Q.~Li, J.~Tang, and D.~Wang, ``Adversarial network embedding,'' in
  \emph{The Thirty-Second {AAAI} Conference on Artificial Intelligence)}, 2018,
  pp. 2167--2174.

\bibitem{jba19}
J.~Sun, B.~Bandyopadhyay, A.~Bashizade, P.~Sadayappan, and S.~Parthasarathy,
  ``{ATP:} directed graph embedding with asymmetric transitivity
  preservation,'' in \emph{The Thirty-Third {AAAI} Conference on Artificial
  Intelligence}, 2019, pp. 265--272.

\bibitem{kplk18}
J.~Kim, H.~Park, J.-E. Lee, and U.~Kang, ``{SIDE}: {R}epresentation learning in
  signed directed networks,'' in \emph{the World Wide Web Conference}, 2018,
  pp. 509--518.

\bibitem{ylzsc15}
C.~Yang, Z.~Liu, D.~Zhao, M.~Sun, and E.~Chang, ``Network representation
  learning with rich text information,'' in \emph{the International Joint
  Conference on Artificial Intelligence}, 2015, pp. 2111--2117.

\bibitem{sv17}
S.~Ganguly and V.~Pudi, ``{Paper2vec}: {Combining} graph and text information
  for scientific paper representation,'' in \emph{the European Conference on
  Information Retrieval}, 2017, pp. 383--395.

\bibitem{smbbgmpa17}
A.~Shatnawi, H.~Mili, G.~E. Boussaidi, A.~Boubaker, Y.-G. Gu{\'e}h{\'e}neuc,
  N.~Moha, J.~Privat, and M.~Abdellatif, ``Analyzing program dependencies in
  {Java} {EE} applications,'' in \emph{the International Conference on Mining
  Software Repositories}, 2017, pp. 64--74.

\bibitem{dmc18}
A.~Decan, T.~Mens, and E.~Constantinou, ``On the impact of security
  vulnerabilities in the npm package dependency network,'' in \emph{the
  International Conference on Mining Software Repositories}, 2018, pp.
  181--191.

\bibitem{msccd13}
T.~Mikolov, I.~Sutskever, K.~Chen, G.~Corrado, and J.~Dean, ``Distributed
  representations of words and phrases and their compositionality,'' in
  \emph{the Neural Information Processing Systems Conference}, 2013.

\bibitem{lm14}
Q.~Le and T.~Mikolov, ``Distributed representations of sentences and
  documents,'' in \emph{the International Conference on Machine Learning},
  vol.~32, no.~2, 2014, pp. 1188--1196.

\bibitem{redhat}
Y.~Fisher, ``{Red Hat Enterprise Linux} builds the foundation for the world's
  fastest supercomputer(s),''
  \url{https://www.redhat.com/en/blog/red-hat-enterprise-linux-builds-foundation-worlds-fastest-supercomputers},
  2018, accessed: 2018-09-05.

\bibitem{w3techs}
M.~Gelbmann, ``Ubuntu became the most popular {Linux} distribution for web
  servers,''
  \url{https://w3techs.com/blog/entry/ubuntu_became_the_most_popular_linux_distribution_for_web_servers},
  2016, accessed: 2018-09-05.

\bibitem{rs10}
R.~{\v R}eh{\r u}{\v r}ek and P.~Sojka, ``\BIBforeignlanguage{English}{Software
  framework for topic modelling with large corpora},'' in
  \emph{\BIBforeignlanguage{English}{the LREC 2010 Workshop on New Challenges
  for NLP Frameworks}}, 2010, pp. 45--50.

\bibitem{kb15}
D.~P. Kingma and J.~L. Ba, ``Adam: {A} method for stochastic optimization,'' in
  \emph{the International Conference on Learning Representations}, 2015.

\bibitem{psm14}
J.~Pennington, R.~Socher, and C.~D. Manning, ``{GloVe}: {G}lobal vectors for
  word representation,'' in \emph{the Conference on Empirical Methods in
  Natural Language Processing}, 2014, pp. 1532--1543.

\bibitem{stackoverflow}
``Developer survey results 2019,''
  \url{https://insights.stackoverflow.com/survey/2019}, 2019.

\end{thebibliography}
\bibliographystyle{IEEEtran}
}

\end{document}